\documentclass[runningheads]{llncs}
\usepackage[T1]{fontenc}
\usepackage{hyperref}
\usepackage{amssymb} 
\usepackage{xcolor}
\usepackage{booktabs} 
\usepackage{graphicx}
\usepackage{algorithm}
\usepackage{algorithmic}
\usepackage{float}
\usepackage{multirow} 
\usepackage{amsmath}
\usepackage{graphicx}
\usepackage{subfig} 

\begin{document}
\title{An Ensemble of Evolutionary Algorithms With Both Crisscross Search and Sparrow Search for Processing Inferior Individuals}
%
%
\author{Mingxuan Du\inst{1}\and
Tingzhang Luo\inst{1}  \and
Ziyang Wang\inst{2}    \and Chengjun Li\thanks{Corresponding author.}\inst{1}      }
\authorrunning{Mingxuan Du et al.}
\titlerunning{EA4eigCS}
\institute{China University of Geosciences, Wuhan \and Jilin University}
%
\maketitle              
\begin{abstract}
In the field of artificial intelligence, real parameter single objective optimization is an important direction.
Both the Differential Evolution (DE) and the Covariance Matrix Adaptation Evolution Strategy (CMA-ES) demonstrate good performance for real parameter single objective optimization.
Nevertheless, there exist other types of evolutionary algorithm for the purpose.
In recent years, researchers begin to study long-term search.
EA4eig - an ensemble of three DE variants and CMA-ES - performs well for long-term search.
In this paper, we introduce two types of evolutionary algorithm proposed recently - crisscross search and sparrow search - into  EA4eig as secondary evolutionary algorithms to process inferior individuals.
Thus, EA4eigCS is obtained.
In our ensemble, the secondary evolutionary algorithms are expected to vary distribution of the population for breaking stagnation. 
Experimental results show that our EA4eigCS outperforms EA4eig and is competitive when compared with state-of-the-art algorithms.  
Code and supplementary material are available at: \url{https://anonymous.4open.science/r/EA4eigCS-2A43}.

\end{abstract}
\section{Introduction}
\label{introduction}
Real parameter single objective optimization for searching the best decision vector in solution space to minimize (or maximize) objective function is an important direction in the field of artificial intelligence.
In recent years, researchers begin to study long-term search in which the maximum number of function evaluations ($MaxFES$), as the termination criterion of algorithm, scales exponentially with the increase of Dimensionality of solution space ($D$).
After all, difficulty of real parameter single objective optimization scales exponentially with the increase of $D$.

So far, types of evolutionary algorithm are used for real parameter single objective optimization.
Among the types, both the Differential Evolution (DE)~\cite{storn1997differential} and the Covariance Matrix Adaptation Evolution Strategy (CMA-ES)~\cite{hansen1996adapting,hansen2003reducing} demonstrate good performance.

In each generation of DE, mutation generates mutant vectors - individuals - based on target vectors. 
Then, based on pairs of mutant vector and target vector, trial vectors are generated by crossover.
In selection, between each pair of trial vector and target vector, the vector better in fitness is selected and then becomes target vector in the next generation.

In each generation of CMA-ES, the individuals better in fitness decide the center of individuals.
Then, offspring are produced based on the center by sampling from a Gaussian distribution.
Among all parents and offspring, individuals betters in fitness are selected for the next generation.

There exist a number of ensembles of DE and CMA-ES.
For example, for long-term search, EA4eig~\cite{bujok2022eigen} is an ensemble of CMA-ES and three DE variants - CoBiDE~\cite{wang2014differential}, IDEbd~\cite{bujok2017enhanced}, and jSO~\cite{brest2017single}.
In the ensemble, for every generation, one of the four constituent algorithms is chosen based on roulette wheel.
The probability of choosing the $h$th algorithm - $p_h$ - is computed as below, 
\begin{equation}
p_h=\frac{n_h+n_0}{\sum_{j=1}^{H}(n_j+n_0)} (h=\{1,2,3,4\}),
\label{eqh1} 
\end{equation}
where $n_h$ represents the success times of the $h$th algorithm in the current generation and $n_0$ is a constant.
In the ensemble, the application range of the covariance matrix learning is extended from CoBiDE to all constituent algorithms.
EA4eig is with the linear population size reduction (LPSR).
The algorithm is the winner of the IEEE Congress on Evolutionary Computation (CEC) 2022 competition on real parameter single objective optimization, the latest CEC competition for long-term search.

EA4eig has more constituent algorithms than most ensembles for real parameter single objective optimization.
For example, also designed for long-term search, both APGSK-IMODE~\cite{mohamed2021gaining} and APGSK-IMODE-FL~\cite{Zhu2023ensemble} have only two constituent algorithms.
In fact, the four constituent algorithms of EA4eig have been proposed for years and were not designed for long-term search at all.
In the framework of EA4eig, search of each constituent algorithm may be interrupted after a generation.
As a result, convergence becomes slow to adapt to long-term search.  

Beside DE and CMA-ES, there exist other types of evolutionary algorithm for real parameter single objective optimization.
Among the types, crisscross search~\cite{meng2014crisscross} and sparrow search~\cite{xue2020novel} were proposed recently.
Although the two types are still in the early stage of research, they do have interesting features.

Crisscross search contains horizontal crossover and vertical one.
The former is executed between pairwise individuals.
All dimensions of the individuals are involved.
However, the latter is carried out among all individuals.
Just two dimensions are considered.

In sparrow search, best individuals in the population are producers, while other ones are scroungers.
The two types of individual are different in behavior.
Moreover, the sparrows at the edge of the group quickly move
toward the safe area to get a better position when
aware of danger, while the sparrows in the middle
of the group randomly walk in order to be close to
others. 

The motivation of this paper is given below.
Now that EA4eig demonstrates good performance for long-term search of real parameter single objective optimization, further improvement on the ensemble is meaningful.
Because of marginal effect, it is not the best choice for improvement that more equally competitive constituent algorithms are added into EA4eig.
Instead, it is worth trying that one or more constituent algorithms are added into framework for processing particular individuals after the roulette wheel.


In this paper, both crisscross search and sparrow search are integrated into EA4eig because of their behavior very distinct from the constituent algorithms in EA4eig.
Thus, we propose EA4eig with both crisscross search and sparrow search - EA4eigCS. 
In our EA4eigCS, provided that the best fitness has not been updated for generations, the worst individuals are processed by crisscross search.
Besides, at the end of all generations, a step in sparrow search is applied for updating the worst individuals.
In our proposed ensemble, CoBiDE, IDEbd, jSO, and CMA-ES are called main constituent algorithms.
Meanwhile, crisscross search and sparrow search are both secondary ones because, in a generation, at most a part of individuals are controlled by the two methods.
Our EA4eigCS may be the ensemble with the largest number of constituent evolutionary algorithms for real parameter single objective optimization. 

We execute experiments based on the CEC 2020, 2021 and 2022 benchmark test suites.
In our experiments, EA4eigCS is compared with state-of-the-art or up-to-date algorithms.
Details of the peers can be found in Section 2.
Results show that our algorithm is competitive.

In summary, we make the following key contributions:

\begin{itemize}
    \item \textit{\textbf{Algorithmic:} }We propose EA4eigCS, an enhanced ensemble of evolutionary algorithms including crisscross search and sparrow search, based on the EA4eig framework.
    \item \textit{\textbf{Methodological:}} We utilize secondary evolutionary algorithms rather than local search for processing inferior individuals rather than the best ones.
    \item \textit{\textbf{Experimental:}} Extensive experiments on CEC benchmarks show that 
    EA4eigCS outperforms state-of-the-art algorithms.
\end{itemize}


\section{Related work}\label{Related}
In this section, firstly, we give more details of crisscross search and sparrow search because our EA4eigCS is based on the two techniques.
Then, we review existing algorithms for long-term search.

\subsection{Crisscross search}
As stated above, crisscross search includes two types of crossover.
In each generation, horizontal crossover is executed firstly.
The $i$th individual in the $g$th generation $\vec x_{i,g}$ and the $i’$th one in the same generation $\vec x_{i',g}$ produce offspring $\vec{MS\_hc}_{i,g}$ and $\vec{MS\_hc}_{i',g}$.
Here,
\begin{equation}
MS\_hc_{j,i,g}=r_1\cdot x_{j,i,g}+(1-r_1)\cdot x_{j,i',g}+c_1\cdot (x_{j,i,g}-x_{j,i',g}),
\label{eqhc} 
\end{equation}
while
\begin{equation}
MS\_hc_{j,i',g}=r_2\cdot x_{j,i',g}+(1-r_2)\cdot x_{j,i,g}+c_2\cdot (x_{j,i',g}-x_{j,i,g}).
\label{eqhc2} 
\end{equation} 
The produced $\vec{MS\_hc}_{i,g}$ and $\vec{MS\_hc}_{i',g}$ compete with $\vec x_{i,g}$ and $\vec x_{i',g}$,
respectively.
After horizontal crossover, vertical crossover is carried out.
The $j1$th dimension of all individuals in the $g$th generation is altered based on crossover with the $j2$th dimension as below,
\begin{equation}
MS\_vc_{j1,i,g}=r\cdot x_{j1,i,g}+(1-r)\cdot x_{j2,i,g}, 
\label{eqvc} 
\end{equation} 
where $i=1,2,...,NP$. 
Similarly, the produced $\vec{MS\_vc}_{i,g}$ ($i=1,2,...,NP$) 
compete with $\vec x_{i,g}$.
In Equations~\ref{eqhc}-\ref{eqvc}, $r1$, $r2$, $r$ are parameters with uniformly distributed random values between $0$ and $1$, while $c1$ and $c2$ are those  with uniformly distributed random values between $-1$ and $1$.

\subsection{Sparrow search}
In sparrow search, all individuals are sorted in descending order of fitness.
The best individuals in the population are producers, while the other individuals are scroungers.
The location of the producer is updated as below,
\begin{equation}
x_{j,i,g+1}=
\left\{
\begin{array}{lr}
x_{j,i,g}\cdot exp(\frac{-i}{\alpha\cdot MaxFES}), & if~R_2<ST,\\
x_{j,i,g}+Q\cdot L, & otherwise,
\end{array}
\right.
\label{eq_pro}
\end{equation}
where $j=1,2,...,D$. 
$R_2\in[0, 1]$ and $ST\in[0.5, 1.0]$) represent the alarm
value and the safety threshold respectively. $Q$ is a random number which obeys normal distribution. $L$ is
a $1\times D$ matrix whose all elements have $1$ in value.
For scroungers, location is updated as follow,
\begin{equation}
x_{j,i,g+1}=
\left\{
\begin{array}{lr}
Q\cdot exp(\frac{x_{j,worst,g}-x_{j,i,g}}{i^2}), & if~i>NP/2,\\
x_{j,best,g}+\left|x_{j,i,g}-x_{j,best,g}\right|\cdot A^+\cdot L, &otherwise,
\end{array}
\right.
\label{eq_s}
\end{equation}
where $A^+=A^T(AA^T)^{-1}$, $A$ is
a $1\times D$ matrix whose all elements have $1$ or $-1$ in value.
Then, all individuals are sorted again.
The first half individuals are ones aware of danger.
Their location is updated as follow,
\begin{equation}
x_{j,i,g+1}=
\left\{
\begin{array}{lr}
x_{j,best,g}+\beta\cdot \left|x_{j,i,g}-x_{j,best,g}\right|, & if~f(x_{j,i,g}))>f(x_{j,best,g}),\\
x_{j,i,g}+K\cdot\frac{\left|x_{j,i,g}-x_{j,worst,g}\right|}{f_i-f_w+\varepsilon}, &if~f(x_{j,i,g}))=f(x_{j,best,g}),
\end{array}
\right.
\label{eq_w}
\end{equation}
where $\beta$ is a normal distribution
of random numbers with a mean value of 0 and a variance of 1, $K\in[-1,1]$ is a random number, $f(\cdot)$ denotes fitness, and $\varepsilon$ is the smallest constant so as to avoid zero-division-error.

\subsection{Existing algorithms for long-term search}
Study of long-term search begins from 2020.
Initially, algorithms for long-term search mainly come from the latest three CEC competitions on real parameter single objective optimization held in 2020, 2021, and 2022, respectively.
Recently, algorithms suitable for long-term search can be found elsewhere.

IMODE~\cite{sallam2020improved}, the winner of the CEC 2020 competition, is a DE variant of three mutation strategies.
The three mutation strategies seize individuals randomly in each generation.
APGSK-IMODE~\cite{mohamed2021gaining}, the winner of non-shifted cases in the CEC 2021 competition, is an ensemble of an adaptive variant of Gaining-Sharing Knowledge (GSK)~\cite{mohamed2020gaining} algorithm and IMODE.
At intervals, the two constituent algorithms share the best individual. 
NL-SHADE-RSP~\cite{stanovov2021nl}, the winner of both shift cases and rotated shifted cases in the same competition, employs 
selective pressure adjustment and automatic tuning of usage probability of the archive for collecting eliminated individuals. 
jDE-21~\cite{brest2021self}, the winner for non-rotated shifted cases, employs a restart mechanism for diversification.
Meanwhile, the crowding mechanism and the scheme to choose individual for mutation proposed in j2020~\cite{brest2020differential} is used in the algorithm after revision.
MLS-L-SHADE~\cite{jou2020multi} contains multiple subpopulations all controlled by the same DE algorithm.
At initial, a given number of individuals are randomly selected as the initial center of subpopulations, respectively.
Then, the number of subpopulations are obtained based on clustering.
Provided that no progress can be made in a subpopulation for given generations, the subpopulation is restarted. 
EA4eig~\cite{bujok2022eigen} has been introduced in Section~\ref{introduction}.
NL-SHADE-LBC~\cite{stanovov2022nl}, the algorithm ranking second in the CEC 2022 competition, is with mutation and crossover modified from those of L-SHADE-RSP~\cite{stanovov2018lshade} and NL-SHADE-RSP.

In \cite{Zhu2023ensemble}, APGSK-IMODE-FL is proposed based on APGSK-IMODE.
In the proposed ensemble, the individuals which have stayed in the population for enough generations and bad in fitness are exchanged between the two constituent algorithms.
In \cite{ye2023differential}, AMCDE is presented based IMODE.
The algorithm is with two states.
In the monopoly state, all individuals are controlled by just one mutation strategy.
In the competition state, all the the three one compete the portion of controlling with each others.

Among the above algorithms, IMODE, APGSK-IMODE, EA4eig, APGSK-IMODE-FL, and AMCDE are all ensemble methods.
IMODE and AMCDE are both DE algorithms with an ensemble of mutation strategies.
APGSK-IMODE and APGSK-IMODE-FL are both ensembles of DE and GSK.
Although EA4eig also belong to ensemble of DE and another type of evolutionary algorithm, compared with APGSK-IMODE and APGSK-IMODE-FL, a much larger number of constituent algorithms are integrated.

\section{Methodology}\label{Methodology}

\begin{algorithm}[h]\scriptsize
	\caption{The pseudo-code of EA4eigCS}
    \label{code}
	\textbf{Input}:\\ $NP_{max}$, the initial value of population size; \\$NP_{min}$, the final value of population size; \\$MaxFES$; \\ $T_{gen}$, the threshold of number of generation without the change in the best fitness,
 \\$R_c$, the ratio of individuals processed by crisscross search; \\ 
 $R_s$, the ratio of individuals considered by sparrow search;\\
	\textbf{Parameter}: \\$NP$, $FES$, and $t$
	\begin{algorithmic}[1]
		\STATE Initialize the initial generation of the population $P_0$
        \STATE Evaluate the $NP_{max}$ individuals in $P_0$
  	    \STATE $NP=NP_{max}$, $FES=NP$, and $t=0$
		\WHILE{$FES<=MaxFES$}
            \STATE Choose an algorithm from the four ones based on probability obtained according to Equation~\ref{eqh1} to control the population
            \STATE $FES=FES+NP$
            \IF{the best fitness is not updated}
                \STATE t=t+1
            \ELSE
                \STATE t=0
            \ENDIF
            \IF{$t=T_{gen}$}
            \STATE Sort individuals in descending order of fitness 
            \STATE Employ crisscross search on the $NP\cdot R_c$ worst individuals
            \STATE $FES=FES+NP\cdot R_c$
            \STATE $t=0$
            \ENDIF
            \STATE Employ sparrow search on the $NP\cdot R_S$ worst individuals
            \STATE $FES=FES+NP$            
        \STATE $NP=\frac{FES}{MaxFES}\cdot (NP_{min}-NP_{max})+NP_{max}$
		\ENDWHILE	
		\STATE {Report solution}
	\end{algorithmic}
\end{algorithm}

Now that EA4eig has good performance, we plan to further improve the algorithm.
For continuous optimization, such as real parameter single objective optimization, stagnation is the main phenomenon demonstrating that no progress in solving can be made any more.
Although EA4eig is with good performance, the ensemble still often falls into stagnation.
Therefore, enhancement requires to be made.
However, it may not be an efficient choice that more constituent algorithms are provided in the roulette wheel beside the original four ones because of marginal effect.
Instead, method should be used for particular individuals in the population for algorithm enhancement.
If one or more individuals are varied by the employed method(s), distribution of population is changed.
Then, stagnation may be avoided. 

In many algorithms for real parameter single objective optimization, local search is used beside the main operators on particular individual for changing distribution of population.
Here, we prefer that evolutionary algorithm different from the four main ones in EA4eig is exerted on particular individuals for further changing distribution of population.
Such evolutionary algorithm added in ensemble is called secondary constituent algorithm in this paper.
Although secondary constituent algorithm may control just a few individuals in a generation, all individuals in the population can participate in its mutation and crossover.
Therefore, compared with in local search, secondary constituent algorithm considers more hereditary information in production of offspring.
In this paper, we select both crisscross search and sparrow search as secondary constituent algorithms.
In our ensemble, the two secondary constituent algorithms are executed just on inferior individual.
Reasons are given below.
Only if offspring produced by the secondary constituent algorithms has better fitness than its parent and then replace the latter, the two methods can exert influence on evolution by varying distribution of population.
It may be difficult for the two methods to improve the current best individuals.
After all, compared with the DE variants and CMA-ES, both crisscross search and sparrow search still remain to be developed.
Nevertheless, the possibility of improving inferior individual is much higher than that of refreshing the best one.
The updating of the worst individuals still leads to change in distribution of population.
In the following execution of the main constituent algorithms, the best fitness may be improved again based on the changed  population. 
More details of integrating crisscross search and sparrow search are given below.

Firstly, provided that the best fitness in the population has not been updated for a given number of generations $T_{gen}$, after the execution of one of the main constituent algorithms, crisscross search is called for the worst $R_c\cdot NP$ individuals in a generation, where $R_c$ is the ratio of individuals processed by crisscross search and $NP$ denotes population size.
The scheme can be explained as below.
The phenomenon that the best fitness in the population has not been updated for a given number of generations demonstrates that, for the main constituent algorithms, improving the current best fitness is difficult.
Therefore, we call crisscross search for helping.

Moreover, at the end of each generation, the step of sparrow search in Equation~\ref{eq_w} is applied on a portion of individuals.
In detail, among the $R_s\cdot NP$ worst individuals, $R_s^2\cdot NP$ ones are randomly chosen for the step, where $R_s$ is the ratio of individuals considered by sparrow search.
That is, in all generations, the step is used to improve individuals among worst ones at a rate. 

Base on the integration of the two search techniques, we realize our EA4eigCS.
We give pseudo-code of EA4eigCS in Algorithm~\ref{code}.
It can be seen from Algorithm~\ref{code} that, compared with EA4eig, our EA4eigCS has the extra three parameters - $T_{gen}$, $R_c$, and $R_s$.

 \section{Experimental study}
\label{exp}
\subsection{Experimental Setup}

\noindent\textbf{Benchmark.} We validate the performance of our algorithm EA4eigCS on two benchmarks - CEC 2021 and 2022 benchmark test suits. 
In our research, we initially conducted a comparative analysis of state-of-the-art algorithms on the CEC 2022 benchmark to identify the most competitive performers. Subsequently, these top-performing algorithms were further evaluated on the CEC 2021 benchmark to confirm their effectiveness. Additionally, the CEC 2022 dataset served as the basis for our ablation studies, enabling us to dissect the contribution of individual components of our proposed method.

\noindent\textbf{Compared methods.}  Beside our EA4eigCS, seven algorithms -  IMODE~\cite{sallam2020improved}, NL-SHADE-RSP~\cite{stanovov2021nl}, APGSK-IMODE~\cite{mohamed2021gaining}, MLS-L-SHADE~\cite{jou2020multi}, EA4eig~\cite{bujok2022eigen}, NL-SHADE-LBC~\cite{stanovov2022nl}, and AMCDE~\cite{ye2023differential} are involved in our experiment as peers. All the peers have been introduced in Section 2. 

\noindent\textbf{Implementation details.} The experimental parameter settings for all algorithms are reported in Table \ref{para}, and all algorithms are run on MATLAB R2021b for fair comparison. We utilize both the Wilcoxon rank sum test and the Friedman test as our evaluation metrics. 
\begin{table}[H]\footnotesize
    \caption{Settings of the involved algorithms}
    \label{para}
    \begin{tabular}{ccc}   
        \toprule
        Algorithm & Parameters & \\\midrule
        \multirow{2}{*}{IMODE} & $NP_{max}=D^2\cdot 6$, $NP_{min}=4$, $A_{rate}=2.60$, \\& $H=D\cdot 20$, $FES_{LS}=MaxFES\cdot 0.85$, and $p=0.3$~\cite{sallam2020improved} \\
        NL-SHADE-RSP & $NP_{max}=30D$, $M_{f,r}=0.2$, $M_{CR,r}=0.2$, and $n_A=0.5$~\cite{stanovov2021nl}\\
        \multirow{2}{*}{APGSK-IMODE} & $NP^{max}_2=\frac{D\cdot30}{4}$, $NP^{min}_2=12$, \\ & $NP^{max}_1=D\cdot 30-NP^{max}_2$, $NP^{min}_1=4$, and $CS=50$ ~\cite{mohamed2021gaining}\\
        MLS-L-SHADE & $NP_{max}=D\cdot 18$, $NP_{min}=4$,$A_{rate}=2.60$, and $H=D^2\cdot 0.36$\\
        EA4eig & $NP_{max}=100$ and $NP_{min}=10$~\cite{bujok2022eigen}\\
        NL-SHADE-LBC & $NP_{max}=D\cdot 20$, $M_{F,r}=0.9$, \\
        &$M_{Cr,r}=0.9$, $k=1$, $\lvert A\rvert = NP \cdot 0.1$, and $n_A=0.5$ ~\cite{stanovov2022nl} \\
        \multirow{3}{*}{AMCDE} & $NP_{max}=D^2\cdot 6$, $NP_{min}=4$, $A_{rate}=2.60$, $H=D\cdot 20$, \\ & $FES_{LS}=MaxFES\cdot 0.85$, $p=0.3$, $G_{n\_init}=5$,  \\ &  $p_{bc1}=0.4$, $p_{bc2}=0.4$, $p_r=0.01$, and $p_w=0.2$~\cite{ye2023differential}   \\        
        EA4eigCS & Beside the parameter setting of EA4eig,\\ 
        &$T_{gen}=3$, $R_c=1/6$, $R_s=1/2$  \\
        \bottomrule
    \end{tabular}
\end{table}

\subsection{Experimental Results}

Firstly, we compare our EA4eigCS with all the peers based on the CEC 2022 benchmark test suite, the latest CEC suite for long-term search.
Results are listed in Tables S1 and S2 of our supplementary material and summarized in Table~\ref{bench2022}. 
As reported in Table~\ref{bench2022}, both Wilcoxon rank sum test and Friedman test demonstrate superiority of our EA4eigCS. 
In terms of Wilcoxon rank sum test, EA4eigCS is even with EA4eig when $D = 10$, while demonstrates better performance in all the other cases.
In terms of Friedman test, our ensemble performs best.

\begin{figure*}[h]  
\centering  
\subfloat[F4]{  
  \includegraphics[width=0.30\textwidth]{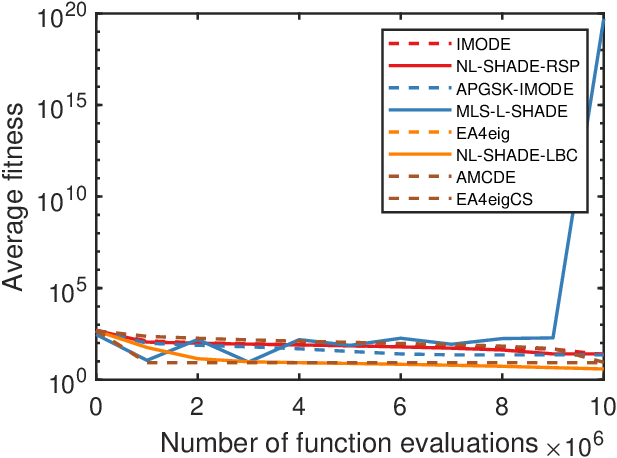}  
  \label{f4}  
}  
\hspace{0.01\textwidth} 
\subfloat[F6]{  
  \includegraphics[width=0.30\textwidth]{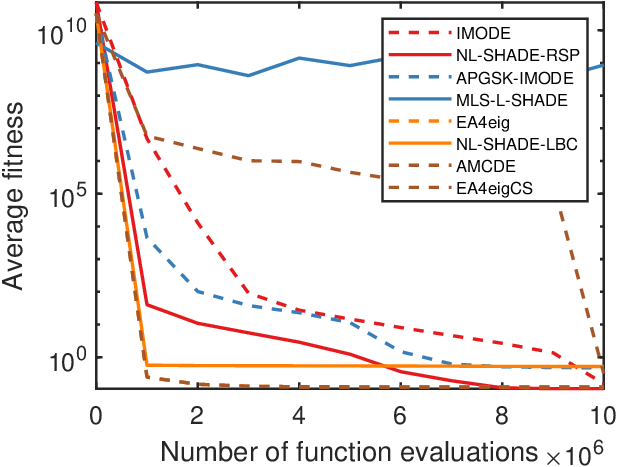}  
  \label{f6}  
}  
\hspace{0.01\textwidth}  
\subfloat[F7]{  
  \includegraphics[width=0.30\textwidth]{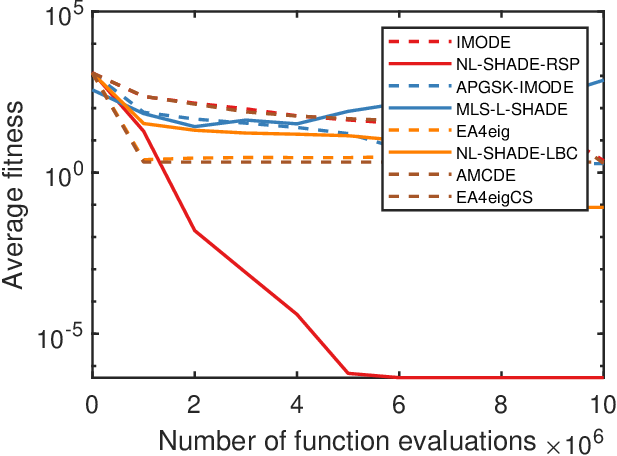}  
  \label{f7}  
}  
  
\bigskip 
  
\subfloat[F8]{  
  \includegraphics[width=0.30\textwidth]{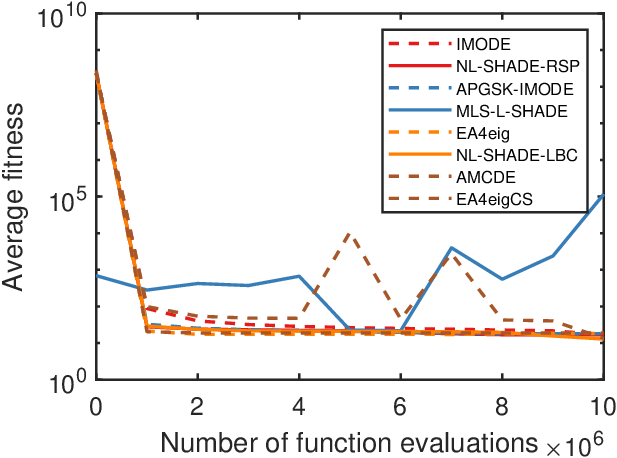}  
  \label{f8}  
}  
\hspace{0.01\textwidth}  
\subfloat[F9]{  
  \includegraphics[width=0.3\textwidth]{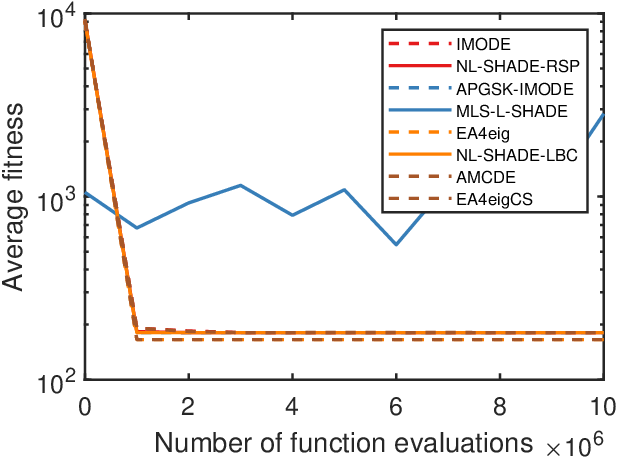}  
  \label{f9}  
}  
\hspace{0.01\textwidth}  
\subfloat[F11]{  
  \includegraphics[width=0.3\textwidth]{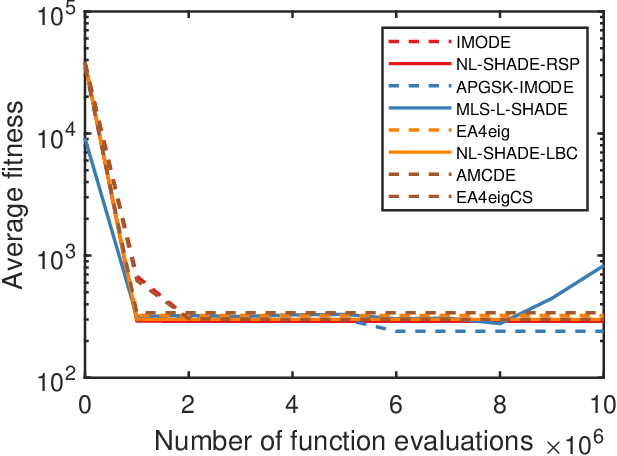}  
  \label{f11}  
}  

\hspace{0.01\textwidth}  
\subfloat[F12]{  
  \includegraphics[width=0.3\textwidth]{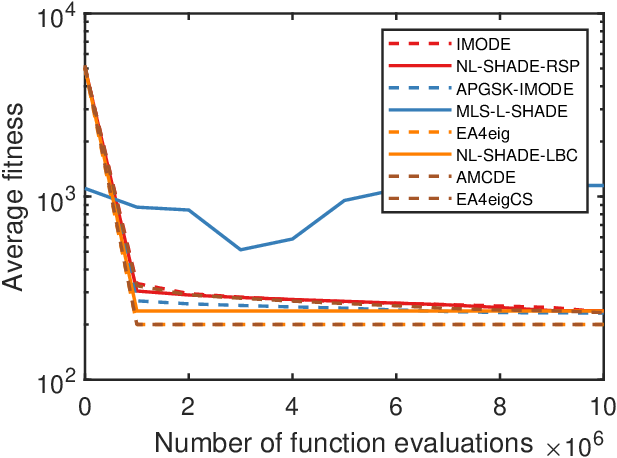}  
  \label{f12}  
}  
  
\caption{Convergence graph of the eight algorithms for seven functions in the CEC 2022 benchmark test suite when $D=20$}  
\label{con1}  
\end{figure*}

We give convergence graph of all the six algorithms for seven functions when $D=20$ in Fig.~\ref{con1}.
The seven functions are chosen because all the algorithms never obtain their global optimal.  For each function, following a certain number of function evaluations (specifically, 11 evaluations), the plot depicts the average fitness value obtained from 30 separate executions.
It can be seen from the figure that the convergence curve of MLS-L-SHADE may go worse during execution.
The phenomenon is attributed to a scheme of subpopulation restart.
The figure demonstrates that, for most of the functions, e.g. F4, F6, F8, F9, F11 and F12, the curve of EA4eigCS and that EA4eig 
is very similar.
In brief, the integration of the two search methods varies little in convergence manner.
\begin{table}[H]
\scriptsize
\centering
\caption{Outcome of the Wilcoxon rank sum test and that of the Friedman test based on results for the CEC 2022 benchmark test suite. "$+$" and "$-$" represent the current algorithm performs significantly better and statistically worse than EA4eigCS. Meanwhile, "$\approx$" means that there is no significant difference}
\label{bench2022}
\setlength{\tabcolsep}{8pt} 
\renewcommand{\arraystretch}{1.2} 
\begin{tabular}{c|c|c|c}
\toprule 
Algorithm &Test type & $D = 10$&  $D = 20$\\
 \hline
\multirow{2}{*}{IMODE~\cite{sallam2020improved}\textcolor{gray}{[CEC'20]}} & Wilcoxon +/-/$\approx$ & 2/4/6 & 1/6/5 \\ \cline{2-4} 
 & Friedman ranking & 4.63 (6) & 5.08 (6) \\ \hline
\multirow{2}{*}{NL-SHADE-RSP~\cite{stanovov2021nl}\textcolor{gray}{[CEC'21]}} & Wilcoxon +/-/$\approx$ & 2/4/6 & 3/5/4 \\ \cline{2-4} 
 & Friedman ranking & 4.50 (5) & 5.08 (6) \\ \hline
\multirow{2}{*}{APGSK-IMODE~\cite{mohamed2021gaining}\textcolor{gray}{[CEC'21]}} & Wilcoxon +/-/$\approx$ & 2/4/6 & 1/9/2 \\ \cline{2-4} 
 & Friedman ranking & 4.21 (3) & 5.46 (8) \\ \hline
\multirow{2}{*}{MLS-L-SHADE~\cite{jou2020multi}\textcolor{gray}{[CEC'20]}} & Wilcoxon +/-/$\approx$ & 3/5/2 & 3/4/5 \\ \cline{2-4} 
 & Friedman ranking & 5.29 (8) & \underline{3.46 (2)} \\ \hline
\multirow{2}{*}{EA4eig~\cite{bujok2022eigen}\textcolor{gray}{[CEC'22]}} & Wilcoxon +/-/$\approx$ & 0/0/12 & 0/2/10 \\ \cline{2-4} 
 & Friedman ranking & 4.21 (3) & 4.83 (5) \\ \hline
\multirow{2}{*}{NL-SHADE-LBC~\cite{stanovov2022nl}\textcolor{gray}{[CEC'22]}} & Wilcoxon +/-/$\approx$ & 0/5/5 & 3/4/5 \\ \cline{2-4} 
 & Friedman ranking & 5.21 (7) & 4.71 (4) \\ \hline
\multirow{2}{*}{AMCDE~\cite{ye2023differential}\textcolor{gray}{[SEC'23]}} & Wilcoxon +/-/$\approx$ & 2/4/6 & 1/6/5 \\ \cline{2-4} 
 & Friedman ranking & \underline{4.17 (2)} & 4.21 (3) \\ \hline
\multirow{2}{*}{EA4eigCS (Ours)} & Wilcoxon +/-/$\approx$ & -/-/- & -/-/- \\ \cline{2-4} 
 & Friedman ranking & \textbf{3.79 (1)} & \textbf{3.17 (1)} \\ \bottomrule 
\end{tabular}
\end{table}
\vspace{-10pt}

\begin{table}[H]
\scriptsize 
\centering
\caption{Outcome of the Wilcoxon rank sum test and that of the Friedman test based on results for the CEC 2021 benchmark test suite}
\label{bench2021}
\setlength{\tabcolsep}{8pt} 
\renewcommand{\arraystretch}{1.2} 
\begin{tabular}{c|c|c|c}
\toprule 
\multirow{2}{*}{Algorithm} & \multirow{2}{*}{Test type} & \multicolumn{2}{c}{2021 Benchmark} \\ \cline{3-4} 
 & & $D = 10$ & $D = 20$ \\ \hline 
\multirow{2}{*}{NL-SHADE-RSP~\cite{stanovov2021nl}\textcolor{gray}{[CEC'21]}} & Wilcoxon +/-/$\approx$ & 4/1/5 & 5/0/5 \\ \cline{2-4} 
 & Friedman ranking & \textbf{2.15 (1)} & \textbf{1.95 (1)} \\ \hline
\multirow{2}{*}{EA4eig~\cite{bujok2022eigen}\textcolor{gray}{[CEC'22]}} & Wilcoxon +/-/$\approx$ & 0/0/10 & 0/0/10 \\ \cline{2-4} 
 & Friedman ranking & 3.70 (5) & 3.25 (3) \\ \hline
\multirow{2}{*}{NL-SHADE-LBC~\cite{stanovov2022nl}\textcolor{gray}{[CEC'22]}} & Wilcoxon +/-/$\approx$ & 0/4/6 & 1/5/4 \\ \cline{2-4} 
 & Friedman ranking & 3.65 (3) & 3.65 (5) \\ \hline
\multirow{2}{*}{AMCDE~\cite{ye2023differential}\textcolor{gray}{[SEC'23]}} & Wilcoxon +/-/$\approx$ & 1/4/5 & 2/4/4 \\ \cline{2-4} 
 & Friedman ranking & 3.65 (3) & 3.40 (4) \\ \hline
\multirow{2}{*}{EA4eigCS (Ours)} & Wilcoxon +/-/$\approx$ & -/-/- & -/-/- \\ \cline{2-4} 
 & Friedman ranking & \underline{3.05 (2)} & \underline{2.75 (2)} \\ \bottomrule
\end{tabular}
\end{table}
\vspace{-12pt}
Then, to further validate the superiority of EA4eigCS, based on the CEC 2021 benchmark, we compare our ensemble with the top-performing peers in the previous comparison. 
Results are listed in Table S3 and S4 of our supplementary material and summarized in Table~\ref{bench2021}. 
According to Table~\ref{bench2021}, when both $D = 10$ and $D = 20$, in terms of Wilcoxon rank sum test, EA4eigCS is comparable to EA4eig and loses to NL-SHADE-RSP, but outperforms the other peers. 
In terms of Wilcoxon rank sum test, our ensemble is defeated by NL-SHADE-RSP, but defeats the other algorithms.

\subsection{Ablation Study}


We conducted an ablation test to evaluate the effectiveness of our schemes. 
We abbreviate sparrow search as SSA and crisscross search as CSO. 
The word - Inferior - is used to represent that both CSO and SSA are just applied to inferior individuals.
The results on the CEC 2022 benchmark test suite with D=20 for four algorithm versions are given in Table S5 and summarized by the outcome of Wilcoxon rank sum test and that of the Friedman test in Table \ref{ablation}. 
It can be seen that both CSO and SSA bring improvement. 
Moreover, when the both methods are integrated, solution is further improved. 
Notably, when CSO and SSA are applied selectively to the predefined inferior individuals rather than all individuals, algorithm becomes even more powerfully.
In brief, each of the proposed steps is crucial. 

\begin{table}[H] 
\centering
\caption{Summary of ablation study}
\label{ablation}
\begin{tabular}{@{}ccc|c|c@{}}
\toprule
SSA & CSO & Inferior  & Wilcoxon +/-/$\approx$&  Friedman ranking\\ \midrule
 &  &  & 0/2/10 & 4.04 (5)\\
 \checkmark &  &  & 0/1/11& 2.54 (3)\\
  & \checkmark &  &0/2/10 & 3.54 (4)\\
 \checkmark & \checkmark &  & 0/0/12 & 2.46 (2)\\
 \checkmark & \checkmark & \checkmark & -/-/-  & 2.42 (1) \\ \bottomrule
\end{tabular}
\end{table}
\section{Conclusion}\label{Conclusion}
Although real parameter single objective optimization has been studied for decades, long-term search begins to be concerned in recent. 
A number of algorithms have been proposed for long-term search.
For example, EA4eig, an ensemble of three DE variants and CMA-ES, is very excellent in performance.

In this paper, we integrate crisscross search and sparrow search into EA4eig as secondary evolutionary algorithms to process inferior individuals and then obtain EA4eigCS.
Experimental results demonstrate that our EA4eigCS is competitive compared with state-of-the-art algorithms for long-term search of real parameter single objective optimization.
Although EA4eig and EA4eigCS are similar in convergence manner, the our proposed ensemble has better performance than the former.

In the future, we may further study ensemble with a large number of constituent evolutionary algorithms for real parameter single objective optimization.
More types of evolutionary algorithms may be considered by us as secondary evolutionary algorithms for building better ensemble.
\bibliographystyle{splncs04}
%
\bibliography{sample-base}

\end{document}